\documentclass[letterpaper]{article} % DO NOT CHANGE THIS
\usepackage[preprint]{aaai2027}  % arXiv preprint: show authors and suppress the conference copyright notice
\usepackage[hyphens]{url}  % DO NOT CHANGE THIS
\usepackage{graphicx} % DO NOT CHANGE THIS
\usepackage{natbib}  % DO NOT CHANGE THIS AND DO NOT ADD ANY OPTIONS TO IT
\usepackage{caption} % DO NOT CHANGE THIS AND DO NOT ADD ANY OPTIONS TO IT
\usepackage{amsmath}
\usepackage{amssymb}
\usepackage{algorithm}
\usepackage{algorithmic}
\usepackage{booktabs}
\usepackage{xspace}
\usepackage{adjustbox}
\usepackage{multirow}

\newcommand{\ours}{\textsc{SelfWAM}\xspace}
\newcommand{\fastwam}{\textsc{FastWAM}\xspace}
\newcommand{\cmark}{\ensuremath{\checkmark}}
\newcommand{\xmark}{\ensuremath{\times}}
\newcommand{\R}{\mathbb{R}}

\newcommand{\Lact}{\mathcal{L}_{\mathrm{act}}}

\title{SelfWAM: A Self-Grounded Unified World Action Model for Fast Robot Control}
\author{
Bikang Pan$^{1,2,*}$ \hspace{1em}
Fan Liu$^{1,2,*}$ \hspace{1em}
Haotao Lu$^{1,2}$ \hspace{1em}
Jingya Wang$^{1}$ \hspace{1em}
Ye Shi$^{1,2}$
}

\affiliations{
$^{1}$ShanghaiTech University \hspace{1em}
$^{2}$InstAdapt\\
$^{*}$Equal contribution.\\
\texttt{\{panbk2023, liufan2025, shiye\}@shanghaitech.edu.cn}\\
\url{https://selfwam.github.io/}
}
\begin{document}

\maketitle

\begin{abstract}
World Action Models (WAMs) improve robot policy learning by jointly modeling actions and future observations. However, conditioning future prediction only on the task prompt and observation context risks capturing generic task progression rather than the action-specific consequences of the executed action. We introduce \ours, a unified self-grounded WAM built on a modality-specialized Mixture-of-Transformers (MoT) architecture that jointly predicts actions, action-conditioned future RGB frames, and robot self-masks, thereby grounding future prediction in the robot's visible body and its action-induced motion. During joint training, \ours allows future visual queries to attend to a clean copy of the demonstrated action, turning the video branch into an action-specific consequence model while leaving the fast action-only inference path unchanged. To focus video learning on action-relevant visual changes, we use prompt-specific objectives for future robot self-mask prediction, which removes appearance details and provides a target whose temporal evolution is tightly coupled with the conditioning action. Together, clean-action conditioning and future self-mask supervision make future predictions more directly reflect how the executed action changes the robot’s visible motion and the surrounding scene.  Experiments on RoboTwin 2.0 and real-world manipulation tasks show that \ours produces more action-sensitive futures and preserves fast policy inference, while improving policy performance.

\end{abstract}

\section{Introduction}

A policy-facing world model should answer not only what is likely to happen next, but also what will change under a particular robot action, where those changes will occur, and how the robot's own body will mediate them. This calls for self-grounded future prediction that explicitly models the robot's visible body and its action-induced motion. This form of grounding is particularly important for manipulation, where the same observation can admit multiple plausible actions, each inducing a different future. World Action Models (WAMs) provide a natural framework for this goal by coupling action prediction with future visual modeling, thereby bringing pretrained video dynamics priors into policy learning.

Recent WAMs exploit pretrained video generators to improve visuomotor policies by jointly learning visual dynamics and robot actions \citep{ye2026dreamzero,ma2026dit4dit}. \fastwam further demonstrates that most of these gains can be realized without future-video generation at inference. It uses future-video prediction only as a co-training objective, while the deployed policy predicts actions directly \citep{yuan2026fastwam}. This design supports efficient closed-loop control. However, in \fastwam, future-video prediction is conditioned on the current observation but not on the demonstrated action that produced the target future. Consequently, the auxiliary objective does not explicitly require the model to distinguish the visual consequences of different actions. This motivates an action-conditioned formulation that explicitly links each target future to the demonstrated action that produced it. 

Action conditioning alone, however, does not necessarily direct the visual objective toward the most action-relevant changes. RGB future prediction must also account for object appearance, texture, lighting, and background variation, much of which is only weakly related to the robot action. These details can dominate the learning objective without providing direct supervision for how the robot moves through the scene. Robot self-mask prediction complements RGB prediction by suppressing appearance-related variation and isolating the robot's visible motion. Because the temporal evolution of the mask is directly determined by the conditioning action, it provides a more focused learning signal for action-specific embodied dynamics. Moreover, the pretrained video backbones underlying WAMs already provide the generative capacity needed to model both RGB and self-mask futures within a unified action-learning framework.

We propose \ours, a unified self-grounded action-conditioned WAM that couples a deployable policy with an action-conditioned world model. \ours retains the modality specialized two stream Mixture-of-Transformers (MoT) architecture of \fastwam, which consists of a pretrained video backbone and a lightweight action expert coupled through mixed attention \citep{yuan2026fastwam}. During training, a clean copy of the demonstrated action is encoded by the action expert and exposed only to future visual queries, allowing the shared video backbone to predict either future RGB observations or robot self-mask videos under separate output prompts. In contrast, the noisy action-prediction queries attend only to the current observation, language instruction, proprioceptive state, and their own noisy action states; they cannot access the clean-action copy or any future visual tokens. This asymmetric information flow prevents target leakage while grounding predicted visual consequences in both the executed action and the robot’s own motion. At deployment, \ours retains current-context encoding and the lightweight action expert while omitting clean-action conditioning and future visual denoising, thereby preserving a fast action-only inference path.

Our contributions are:
\begin{itemize}
    \item We extend the two-stream MoT with a conditioning path for a clean copy of the demonstrated action, unifying action prediction and action-conditioned future modeling within a single architecture. By exposing this action condition only to future visual tokens, the model can distinguish the future consequences induced by different actions while preserving the lightweight action-only inference path.
    \item Building on this unified design, we introduce future robot self-mask sequence prediction as an auxiliary objective. By excluding appearance and background variation, self-mask prediction focuses action-conditioned modeling on the robot's future visible motion and strengthens the self-grounding of future prediction.
    \item Experiments on RoboTwin 2.0 and real-world manipulation tasks show that \ours improves future prediction fidelity and sensitivity to action changes, strengthens policy performance, and keeps the action-only inference cost nearly unchanged.
\end{itemize}

\begin{figure*}[t]
\centering
\includegraphics[width=0.96\textwidth]{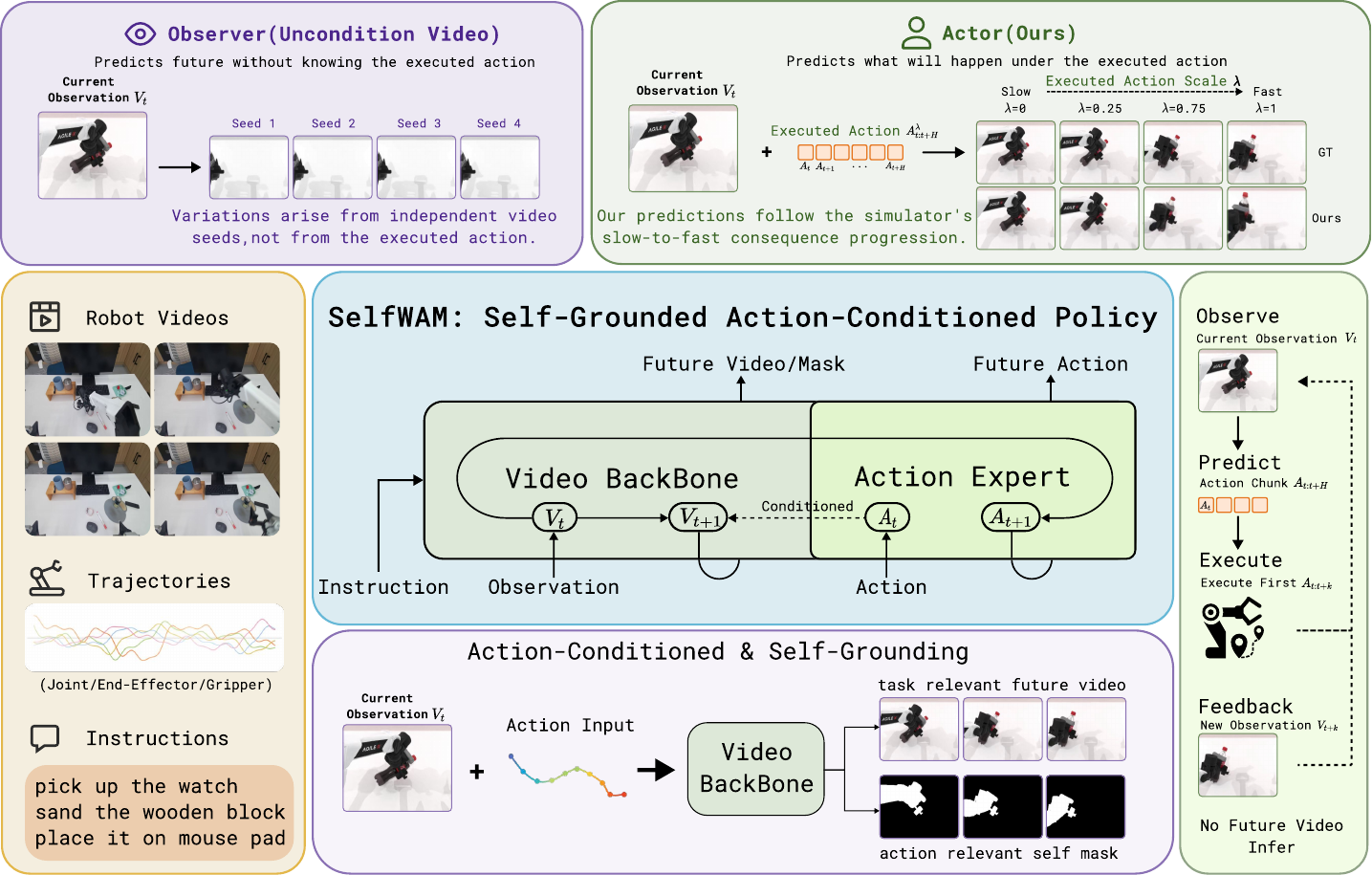}

\caption{\textbf{Overview of SelfWAM.} Compared with action-unconditioned video prediction, SelfWAM shifts the model from a passive observer to an actor that predicts the visual consequences of its own actions, making future generation both action-sensitive and action-consistent. As shown in the top comparison, its predicted motion follows the same slow-to-fast progression as the ground truth when the executed action is scaled. Given robot videos, trajectories, and instructions, it couples a lightweight action expert with a shared video backbone. During training, the executed action conditions a shared video backbone that is trained
on both future RGB and future self-mask prediction through separate prompts, aligning visual dynamics with control while directing attention toward action-induced robot motion rather than background appearance and texture. At deployment, future-video denoising and mask generation are omitted.
The current observation is encoded once, after which iterative action
denoising proceeds only through the lightweight action expert.}
\label{fig:teaser}
\end{figure*}

\section{Related Work}

\subsection{World-Action Models}

Modern robot policies directly map visual observations, language, and proprioception to continuous actions, ranging from task-specific diffusion policies to large-scale generalist vision-language-action models \citep{chi2023diffusionpolicy,kim2024openvla,octo2024,2024pi_0,2025pi_05}. Although effective for action prediction, their training objectives do not explicitly require modeling the visual consequences of a selected action.

Earlier unified formulations establish that video and action prediction can be learned within a shared generative model. UVA learns joint video-action representations with decoupled output heads, allowing policy inference to bypass video generation, while UWM uses modality-specific diffusion timesteps to represent policies, forward and inverse dynamics, and video prediction within a single model \citep{li2025uva,zhu2025uwm}.

More recent World-Action Models integrate future visual prediction more directly into policy learning. DreamZero, LingBot-VA, DiT4DiT, and ImageWAM jointly generate actions and visual futures or use video representations to support action prediction \citep{ye2026dreamzero,2026lingbotva,ma2026dit4dit,zhang2026imagewam}. \fastwam instead uses future video prediction only as a training objective, enabling direct action inference without future visual generation \citep{yuan2026fastwam}. GigaWorld-Policy instead makes future prediction action-conditioned within a shared Transformer while retaining optional visual rollout at deployment \citep{ye2026gigaworldpolicy}. $\tau_0$-WM further uses action-conditioned futures to evaluate and refine candidate actions at test time \citep{zhou2026tau0wm}.

\ours retains the efficient policy interface of \fastwam but makes future prediction explicitly dependent on the demonstrated action. A clean action copy is exposed only to future visual queries, preventing target leakage into action prediction. Beyond RGB prediction, \ours predicts future robot self masks to provide embodiment-focused supervision without changing the deployed action path.

\subsection{Structured Visual and Robot Body Modeling}

Structured visual targets can focus predictive learning on geometry, semantics, or task-relevant entities. GeoSem-WAM augments RGB prediction with future geometry and semantic supervision, while Mask World Model and MaskWAM use semantic or object masks as predictive targets or prompts \citep{2026geosem,lou2026maskworldmodel,yu2026maskwam}. These representations primarily describe scene structure or task objects. In contrast, \ours predicts future robot self masks to directly supervise action-induced motion of the visible robot body.

Prior work on robot self-recognition has segmented robot hands, learned visual body models, and associated visual observations with proprioceptive signals \citep{almeida2021hand,chen2022selfmodel,chen2026selfother}. Robot masks have also been used to reduce appearance differences across embodiments \citep{lepert2025shadow}. These studies motivate explicit robot-centric representations, but do not integrate action-conditioned future RGB and self-mask prediction with an independently deployable policy.

\section{SelfWAM}

\subsection{Problem Formulation}

At control step $t$, the current context is defined as
\begin{equation}
 c_t=(o_t,q_t,\ell_t),
\end{equation}
where $o_t$ is the current multi-view RGB observation, $q_t$ is proprioception, and $\ell_t$ is the language instruction. The target is an $H$-step action chunk with action dimension $d_a$
\begin{equation}
 \mathbf{a}_t=(a_t,\ldots,a_{t+H-1})\in\R^{H\times d_a}.
\end{equation}
For world-model supervision, we use the next $K$ consecutive frames within the temporal horizon covered by the action chunk:
\begin{equation}
\begin{aligned}
 \mathbf{o}_t^{\mathrm{fut}} &= (o_{t+1},\ldots,o_{t+K}).
\end{aligned}
\end{equation}
The optimal policy output is defined as the action chunk that minimizes the conditional expected action prediction loss:
\begin{equation}
 \pi^*(c_t)=\arg\min_{\hat{\mathbf a}}
 \mathbb E\!\left[
 \ell_a(\hat{\mathbf a},\mathbf a_t)\mid c_t
 \right].
\label{eq:policy}
\end{equation}
Similarly, the optimal world-model output is defined as the future observation sequence that minimizes the conditional expected visual-prediction loss given the current context and the conditioning action:
\begin{equation}
 F^*(c_t,\mathbf a_t)=\arg\min_{\hat{\mathbf o}}
 \mathbb E\!\left[
 d_v(\hat{\mathbf o},\mathbf o_t^{\mathrm{fut}})\mid c_t,\mathbf a_t
 \right].
\label{eq:world}
\end{equation}

Thus, the policy branch learns which action is best supported by the demonstrations for the current observation, while the world branch learns which future observation is most consistent with the observation-action pair. 

\subsection{Architecture}

\begin{figure*}[t]
\centering
\includegraphics[width=0.96\textwidth]{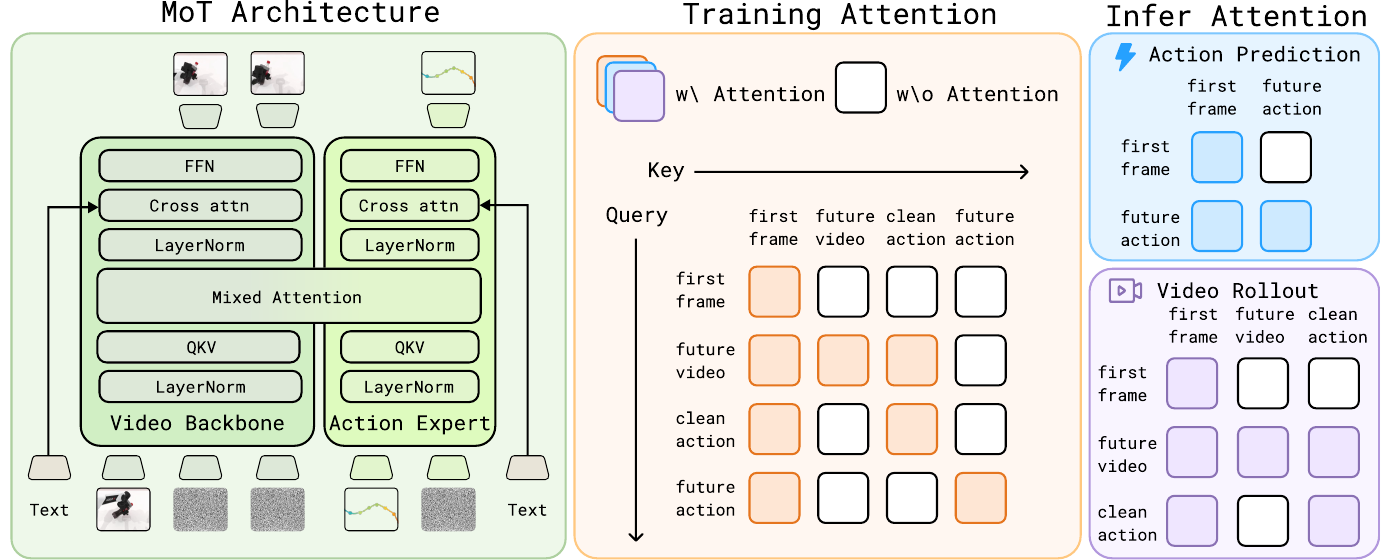}
\caption{\textbf{SelfWAM architecture and attention design.} Left: the modality-specialized MoT contains a video backbone and a lightweight action expert coupled through mixed attention. Middle: during training, clean executed-action tokens condition only future-video queries, aligning the predicted dynamics with the executed control while preventing ground-truth action leakage into future-action prediction. This keeps the policy path consistent with action-only deployment. Right: SelfWAM supports two inference modes. For fast control, the action expert predicts actions directly from the current observation without generating future video. For optional world-model rollout, a supplied action and output prompt condition the video backbone to generate either future RGB or future self-mask video.}
\label{fig:architecture}
\end{figure*}

\ours builds on the two-stream MoT design shown in Fig.~\ref{fig:architecture}. The video stream uses a pretrained video diffusion backbone to jointly process the encoded current observation and noisy future-video latents. The action stream contains noisy future-action tokens processed by a lightweight action expert. All parameters of the action expert are initialized by interpolating the corresponding weights from the pretrained video expert to match the action-stream architecture. The two experts retain modality-specific projections and feed-forward layers, while mixed attention controls which token groups may exchange information. After mixed attention, the tokens are routed back to their respective modality streams and processed by the corresponding modality-specific feed-forward networks. Language and proprioceptive features condition both streams.

The architecture produces future actions through the action expert and future visual targets through the video backbone. The visual target can be either RGB video or robot self-mask video, as described below. The video stream is used only for co-training and optional rollout. Action-only inference retains the current-observation encoding and the lightweight action expert.

\subsection{Information Flow}

In the original \fastwam information flow, the action and video streams are co-trained but remain action-agnostic on the world-model side: noisy future-action queries attend to the current context and noisy action sequence, whereas future-video queries attend only to the current context and their own video tokens \citep{yuan2026fastwam}. Consequently, the video stream is not explicitly informed of the action that produced the demonstrated future. \ours preserves the original policy-side dependency and introduces one additional one-way route: the executed action is exposed to future-video queries while remaining hidden from noisy future-action queries.

To instantiate this route, the standard action stream receives a corrupted future-action chunk and learns to denoise it from the current context. We create a second copy of the same action chunk as a clean conditioning stream. This copy is kept uncorrupted, encoded by the same action tokenizer and action blocks, and assigned the clean denoising timestep $t=0$. At action index $i$, the clean token uses the same temporal position encoding as the corresponding noisy future-action token. The two streams therefore share action semantics and horizon alignment, while their timestep embeddings distinguish a clean condition from a denoising state.

Let $\mathcal{O}$ denote the current-context tokens, $\widetilde{\mathcal{A}}$ the noisy future-action tokens, $\mathcal{A}$ the clean-action tokens, and $\widetilde{\mathcal{V}}$ the noisy future-visual tokens. For a query group $\mathcal{Q}_{X}$, we write $\mathcal{Q}_{X}\rightarrow\mathcal{S}$ to indicate that its queries may attend to the token groups in $\mathcal{S}$:
\begin{equation}
\begin{aligned}
\mathcal{Q}_{\widetilde{\mathcal A}}
&\rightarrow
\{\mathcal O,\widetilde{\mathcal A}\}
\vphantom{\widetilde{\mathcal V}},\\[2pt]
\mathcal{Q}_{\widetilde{\mathcal V}}
&\rightarrow
\{\mathcal O,\widetilde{\mathcal V},\mathcal A\}
\vphantom{\widetilde{\mathcal V}},\\[2pt]
\mathcal{Q}_{\mathcal A}
&\rightarrow
\{\mathcal O,\mathcal A\}
\vphantom{\widetilde{\mathcal V}}.
\end{aligned}
\end{equation}

Thus, future-video prediction is explicitly tied to the demonstrated action, while the policy cannot recover its target from either the clean-action copy or future-video representations. The policy-side computation graph remains identical to action-only deployment.

The clean-action stream serves only as a world-model condition rather than a second action-prediction target. It is enabled during world-model co-training and optional visual rollout, but removed during action-only inference. For rollout, a policy sample or any candidate action can be inserted into the clean-action slots using the same $t=0$ and temporal-position convention, turning the video stream into an action-conditioned world model.

\subsection{Robot Mask Prediction}

We reuse the same video backbone for RGB prediction and robot self-mask prediction. Each demonstration yields two world-model training instances with identical current RGB observation, instruction, proprioception, and clean-action condition. They differ only in the output prompt and the future target. The RGB prompt selects the future RGB clip, while the self-mask prompt selects the aligned robot-segmentation clip. Both output modes use the same spatiotemporal token layout, noise schedule, position encodings, and directed attention pattern.

This formulation treats self-mask prediction as a second output domain of the video model rather than a separate expert. The observation remains RGB in both cases, so the model must infer the robot body from the same visual context used by the policy. The action condition is also unchanged, requiring the generated mask sequence to follow the demonstrated robot motion.

The mask-prompted instance does not supervise action prediction. It contains the clean action only as a condition for the video backbone; no noisy future-action target or action-denoising loss is added for this duplicated instance. The policy target is therefore counted once, while RGB and mask prediction provide two complementary world-model objectives.

\subsection{Training and Inference}

We use the standard flow-matching objectives of the base action and video models. Let $\Lact$ denote the action-denoising loss, and let $\mathcal L_{\mathrm{rgb}}$ and $\mathcal L_{\mathrm{mask}}$ denote the video-denoising losses under the RGB and self-mask prompts, respectively. We define $\mathcal L_{\mathrm{rgb/mask}}$ as their mixture expectation. The overall training objective is
\begin{equation}
 \mathcal L
 = \lambda_{\mathrm{act}}\Lact
 + \lambda_{\mathrm{video}}\mathcal L_{\mathrm{rgb/mask}}.
\label{eq:total}
\end{equation}
The action loss is evaluated once for each policy-training example. The mask-prompted copy contributes only $\mathcal L_{\mathrm{mask}}$.

At deployment, \ours follows the action-only path: it encodes the current context, denoises an action chunk with the lightweight action expert, executes the first actions, and replans. Clean-action and future-video tokens are not instantiated. For optional diagnosis or counterfactual rollout, a candidate action is inserted through the clean-action path and the output prompt selects either a future RGB video or a future self-mask video.
\begin{figure*}[htbp]
\centering
\includegraphics[width=\textwidth]{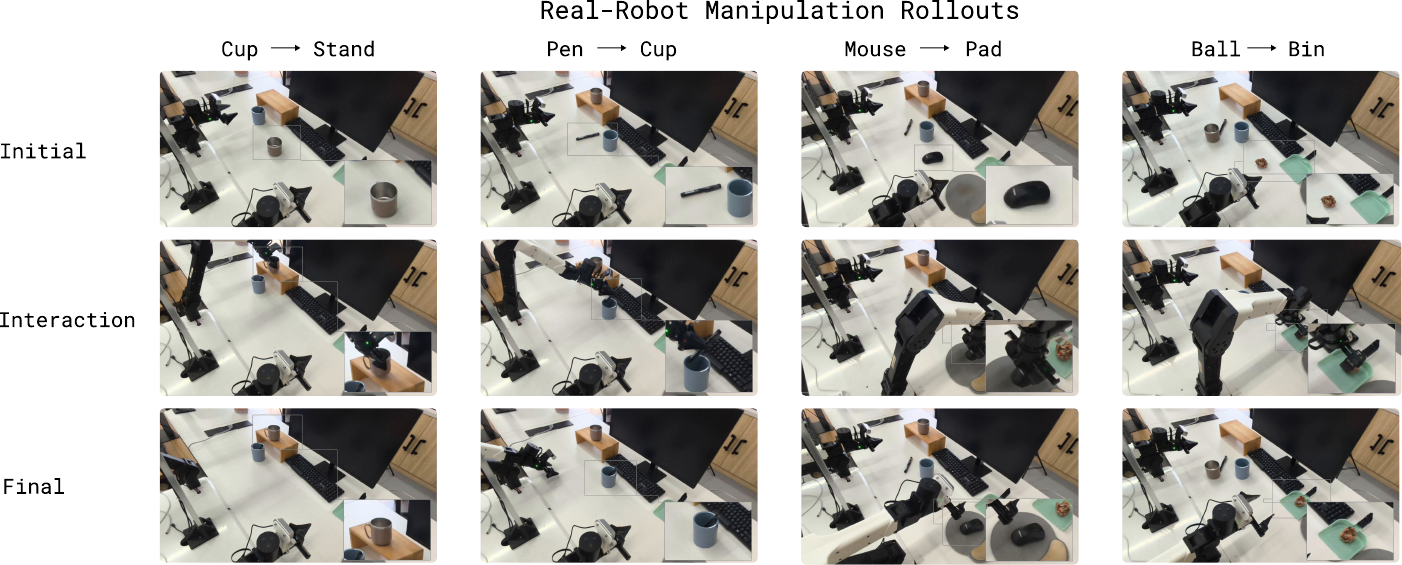}
\caption{\textbf{Representative real-robot manipulation rollouts.}
Each column shows a closed-loop execution of \ours on one of four tasks:
Cup $\rightarrow$ Stand, Pen $\rightarrow$ Cup, Mouse $\rightarrow$ Pad,
and Ball $\rightarrow$ Bin. The rows depict the initial scene, the main
interaction stage, and the final outcome, respectively. Zoomed-in insets
highlight the manipulated object and the corresponding target region.}
\label{fig:real_world}
\end{figure*}
\section{Experiments}

Our experiments span RoboTwin 2.0 and real-world manipulation settings, with the goal of evaluating policy performance, action-grounded future prediction, and deployment efficiency. We further analyze the contribution of each component through controlled ablations and action perturbation studies.

\subsection{Implementation Details}

We initialize the shared video backbone from WAN2.2-5B \citep{wan2025}. The lightweight action expert is initialized from the same WAN2.2-5B checkpoint through parameter interpolation. The policy receives three camera views, which are arranged into a T-shaped composite image with a resolution of $384\times320$. The action horizon is 32 steps, and the corresponding 32-frame future video sequence is temporally downsampled to 8 frames for visual prediction.

We assign equal weights to the action and visual flow-matching objectives, i.e., $\lambda_{\mathrm{act}}=1,\lambda_{\mathrm{video}}=1$. RGB-prompted and self-mask-prompted visual examples are sampled at a ratio of $9:1$.  We train all models for 5 epochs using AdamW with a learning rate of $1\times10^{-4}$. All training runs are conducted on Ubuntu 22.04 with 1 TB of system memory, using Python 3.10.12 and PyTorch 2.7.1 compiled with CUDA 12.8. Training is distributed across 64 NVIDIA A800 GPUs with a global batch size of 1024, corresponding to a per-GPU batch size of 16, and takes approximately 30 hours.

At inference, the policy uses 10 denoising steps to predict a 32-step action chunk. It executes the first 24 actions before replanning from the latest observation.

\subsection{Benchmarks}

We conduct experiments in RoboTwin 2.0 and on a physical bimanual robot. Together, these benchmarks measure policy performance in controlled simulation and physical environments, with task success as the primary metric and inference latency reported to characterize deployment efficiency.

\paragraph{RoboTwin 2.0.}
RoboTwin 2.0 is a large-scale benchmark for coordinated bimanual manipulation, covering more than 50 diverse tasks \citep{chen2025robotwin}.  Following the multi-task protocols adopted in prior work \citep{bi2026motus, yuan2026fastwam}, we train each model on a combined dataset consisting of 2,500 demonstrations from clean environments and 25,000 demonstrations collected with extensive scene randomization. We evaluate performance separately in clean and randomized settings and report the mean task success rate over 100 rollouts per task. The simulator provides exact per-camera robot masks, enabling controlled evaluation of body motion and background randomization.

\paragraph{Real robot.}
We evaluate \ours on the AgileX ALOHA dual-arm platform. The training dataset covers diverse tabletop manipulation tasks and contains approximately 17K episodes and 9.5M transitions. For comparison, all compared methods are trained on the same dataset. For real-world self-mask supervision, we apply RoboSeg \citep{mei2026robotseg} offline to the collected RGB observations, producing robot-body masks that serve as future self-mask targets during training.

\subsection{Policy Performance}

\paragraph{Baselines.}
We compare \ours with representative VLA and WAM baselines. The VLA baselines include $\pi_0$ \citep{2024pi_0} and $\pi_{0.5}$ \citep{2025pi_05}, two strong generalist visuomotor policies. We also include three recent WAMs, Motus \citep{bi2026motus}, GigaWorld-Policy \citep{ye2026gigaworldpolicy}, and \fastwam \citep{yuan2026fastwam}, which jointly learn robot actions and future visual dynamics.
\begin{table}[t]
\centering
\caption{RoboTwin 2.0 success rates (\%).}
\label{tab:main}
% \scriptsize
\resizebox{0.85\columnwidth}{!}{%
\begin{tabular}{@{}lccc@{}}
\toprule
Method & Clean & Random & Average \\
\midrule
$\pi_0$               & 65.92 & 58.40 & 62.16 \\
$\pi_{0.5}$           & 82.74 & 76.76 & 79.75 \\
Motus                 & 88.66 & 87.02 & 87.84 \\
GigaWorld-Policy      & 86.36 & 85.04 & 85.70 \\
\fastwam              & 91.82 & 91.86 & 91.84 \\
\textbf{\ours}        & \textbf{92.16} & \textbf{93.08} & \textbf{92.62} \\
\bottomrule
\end{tabular}%
}
\end{table}

\paragraph{RoboTwin 2.0 evaluation.}
Table~\ref{tab:main} reports policy performance on RoboTwin 2.0. \ours achieves performance comparable to or better than all baselines in both clean and randomized settings. These results show that introducing action-conditioned future prediction and robot self-grounding does not compromise policy performance, while preserving the lightweight action-only inference path. The improvement under visual randomization further suggests that self-grounded supervision helps maintain robustness to appearance variation.

\paragraph{Real-robot evaluation.}
We further evaluate \ours on four real-world pick-and-place tasks: placing a brown cup on a wooden stand, inserting a black pen into a blue cup, placing a black mouse on a gray mouse pad, and depositing a paper ball into a desktop bin. These tasks cover both precise placement onto a target surface and insertion into a receptacle. We compare \ours against \fastwam and $\pi_{0.5}$ under the same initial-state distribution and evaluation protocol. Each method is evaluated for 10 trials per task, and a trial is considered successful only when the target object is grasped and placed in the specified destination without human intervention.

\begin{table}[t]
\centering
\caption{
Success rates (\%) on four real-world manipulation tasks.
}
\label{tab:real_robot_results}

\begingroup
\setlength{\tabcolsep}{4.5pt}
\small

\begin{tabular}{@{}lccc@{}}
\toprule
Task
& \textbf{SelfWAM}
& FastWAM
& $\pi_{0.5}$ \\
\midrule
Brown Cup on Wooden Stand
    & 100 & 80 & 90 \\
Black Pen in Blue Cup
    & 90 & 90 & 90 \\
Black Mouse on Gray Mouse Pad
    & 100 & 80 & 90 \\
Paper Ball in Desktop Bin
    & 90 & 90 & 90 \\
\midrule
Average
    & \textbf{95} & 85 & 90 \\
\bottomrule
\end{tabular}

\endgroup
\end{table}

As shown in Table~\ref{tab:real_robot_results}, \ours achieves policy performance comparable to or better than the strong baselines across all four real-world tasks. These results indicate that the proposed world-model objectives transfer effectively to physical deployment without compromising closed-loop control performance. Figure~\ref{fig:real_world} presents representative rollouts.

% \subsection{Metrics}

% \paragraph{Policy.}
% We report task success under clean and randomized RoboTwin settings and success on real-world manipulation tasks. Closed-loop success is primary.

% \paragraph{Future video.}
% We report LPIPS, PSNR, and FVD-I3D \citep{zhang2018lpips,unterthiner2018fvd}. These metrics measure visual quality but do not by themselves establish action use.

% \paragraph{Efficiency.}
% We report end-to-end latency and peak GPU memory separately for action-only and action-plus-video inference.

\begin{figure*}[t]
\centering
\includegraphics[width=0.9\textwidth]{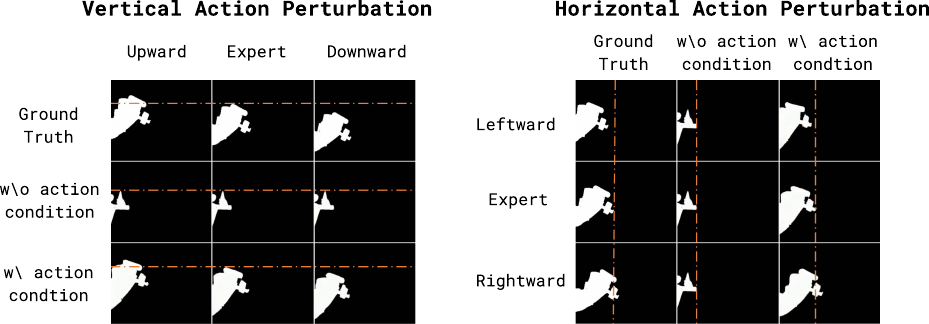}
\caption{\textbf{Directional action perturbation.} We perturb the demonstrated
action along the vertical and horizontal translation axes. The
action-unconditioned baseline remains largely invariant across action variants,
whereas \ours moves the predicted robot self-mask in the commanded direction
and follows the corresponding ground-truth trend. Orange dashed lines provide
fixed spatial references.}
\label{fig:action_perturbation}
\end{figure*}

\subsection{Action-Conditioned Future Prediction}

\paragraph{Future prediction evaluation.}
We next examine the effect of demonstrated-action conditioning on future prediction. We uniformly sample 275 trajectories, corresponding to 1\% of the 27,500 demonstrations in the RoboTwin 2.0 multi-task dataset, and evaluate future prediction over the full duration of each trajectory. Both models are evaluated on the same trajectory subset, differing only in whether future prediction is conditioned on the observation context alone or additionally on the clean demonstrated action. We evaluate future-video fidelity using LPIPS~\citep{zhang2018lpips}, PSNR~\citep{huynh2008scope}, and FVD with I3D features~\citep{unterthiner2018towards}.

\begin{table}[t]
\centering
\caption{Future-video quality on 275 trajectories.}
\label{tab:videoquality}
\resizebox{0.9\columnwidth}{!}{%
\begin{tabular}{@{}lccc@{}}
\toprule
Model &  LPIPS$\downarrow$ & PSNR$\uparrow$ & FVD-I3D$\downarrow$ \\
\midrule
\fastwam (uncond.) & 0.0636 & 29.24 & 45.92 \\
\ours & \textbf{0.0429} & \textbf{32.79} & \textbf{34.55} \\
\bottomrule
\end{tabular}%
}
\end{table}

Compared with the observation-conditioned \fastwam objective, action conditioning yields lower LPIPS and FVD-I3D together with higher PSNR, indicating improved perceptual similarity, pixel-level fidelity, and overall video-distribution quality. This improvement primarily arises from conditioning future prediction on the controllable action sequence aligned with each offline trajectory, allowing the model to generate consequences that more closely match the demonstrated future.

\paragraph{Directional action perturbation.}
To construct perturbed action conditions, we apply a joint-space offset that ramps from $0$ to $\pm 0.10\,\mathrm{rad}$ over the first eight steps and remains fixed thereafter. Directional labels are determined by the induced robot-arm displacement in the high-camera image plane, yielding upward, downward, leftward, and rightward variants. For each variant, we keep the observation, instruction, and sampling noise fixed to isolate the effect of action perturbation. Figure~\ref{fig:action_perturbation} compares the resulting self-mask predictions with the corresponding ground-truth motion trends.

\subsection{Component Ablation}

Table~\ref{tab:robotwin_ablation} examines the incremental contributions of clean-action conditioning and future self-mask supervision. Introducing action-conditioned future RGB prediction alone largely preserves the policy performance of \fastwam, suggesting that the modified information flow does not substantially interfere with action learning. Adding future self-mask supervision then improves performance in both clean and randomized settings, leading to the strongest overall results. This indicates that clean-action conditioning can enrich the world-modeling objective without severely compromising control, while self-mask prediction provides a more embodiment-focused training signal that better supports policy learning.

\begin{table}[H]
\centering
\caption{Component ablation on RoboTwin 2.0. Action Cond. denotes clean-action conditioning for future-RGB prediction, and Self-Mask denotes future robot self-mask supervision. Success rates are reported in \%.}
\label{tab:robotwin_ablation}
\resizebox{\columnwidth}{!}{%
\begin{tabular}{@{}lccccc@{}}
\toprule
Method & Action Cond. & Self-Mask & Clean & Random & Average \\
\midrule
\fastwam
& \xmark & \xmark
& 91.82 & 91.86 & 91.84 \\
Ours w/o self-mask
& \cmark & \xmark
& 90.80 & 90.80 & 90.80 \\
\textbf{\ours}
& \cmark & \cmark
& \textbf{92.16} & \textbf{93.08} & \textbf{92.62} \\
\bottomrule
\end{tabular}%
}
\end{table}

\subsection{Inference Efficiency}

We evaluate inference efficiency on a single NVIDIA H200 GPU using a fixed episode, a 32-step action prediction horizon, and 10 denoising steps. Action-only inference encodes the current context once and omits the clean-action and future-visual streams; the joint setting additionally performs action-conditioned future-RGB denoising, but not future self-mask generation.

\begin{table}[t]
\centering
\caption{Inference efficiency after warm-up, averaged over four runs. Relative changes are computed from the unrounded timings.}
\label{tab:efficiency}
\resizebox{0.9\linewidth}{!}{%
\begin{tabular}{@{}l|cc|c@{}}
\toprule
Measurement & \fastwam & \ours & Change \\
\midrule
Action only (ms) & 320.7 & 323.9 & +0.97\% \\
Action + video (ms) & 668.4 & 677.5 & +1.36\% \\
Peak memory (GiB) & 13.959 & 13.959 & 0.00\% \\
\bottomrule
\end{tabular}%
}
\end{table}

\ours increases action-only latency by only 0.97\% and joint action and video latency by 1.36\%, with no measured increase in peak memory. The auxiliary clean-action and future-visual streams therefore add world-model capability without materially changing the deployment cost of the iterative policy path.

\section{Conclusion}

We introduced \ours, a self-grounded World Action Model that turns future prediction from a generic auxiliary objective into action-conditioned consequence modeling. Its asymmetric conditioning scheme allows the visual branch to learn futures associated with the demonstrated action while keeping policy learning isolated from clean target actions and future observations. Future self-mask supervision further focuses this predictive objective on the robot's visible motion, providing an embodiment-specific signal complementary to RGB prediction. Across RoboTwin 2.0 and real-world manipulation tasks, \ours improves policy performance, future-video fidelity, and sensitivity to controlled action perturbations, while retaining fast action-only inference.

\section*{Acknowledgements}

This work was supported by the National Natural Science Foundation of China under Grants 62406195, the HPC Platform of ShanghaiTech University, and Key Laboratory of Intelligent Perception and Human-Machine Collaboration (ShanghaiTech University), Ministry of Education. Computational resources were also provided in part by Fcloud Co., Ltd.

% Append the supplementary material to the same arXiv PDF.
% This file is included by SelfWAM_AAAI2027_Draft.tex.
% It intentionally has no \documentclass, preamble, bibliography, or \end{document}.

\clearpage
\twocolumn[
\begin{center}
{\LARGE\bfseries Appendix}\par
\end{center}
\vspace{0.8em}
]
\appendix
\setcounter{figure}{0}
\renewcommand{\thefigure}{S\arabic{figure}}
\setcounter{table}{0}
\renewcommand{\thetable}{S\arabic{table}}
\setcounter{equation}{0}
\renewcommand{\theequation}{S\arabic{equation}}

This supplementary material provides implementation and reproducibility details, task-level simulation results, controlled analyses of action sensitivity, diagnostics of the clean and noisy action paths, and the real robot protocol. 

\section{Implementation and Reproducibility Details}
\label{sec:supp_implementation}

This section specifies how training instances are assembled, how action and visual targets are aligned, and which optimization and inference settings are used. Architectural definitions and the directed attention pattern are presented in the main paper and are not repeated here.

\subsection{Training Instance Construction}
\label{sec:supp_training_instances}

Each demonstration window is used to construct either an RGB training instance or a self-mask training instance. Both instance types share the current RGB observation, language instruction, proprioceptive state, clean action condition, video noise schedule, and spatiotemporal positions. They differ in the video prompt, visual target, and action supervision, as summarized in Table~\ref{tab:supp_instance_construction}.

\begin{table}[!ht]
\centering
\caption{Construction of the two training instance types. Both use the clean action as a condition for future visual prediction, while only the RGB instance receives action supervision.}
\label{tab:supp_instance_construction}
\setlength{\tabcolsep}{4pt}

\resizebox{\columnwidth}{!}{
\begin{tabular}{@{}lcccl@{}}
\toprule
Instance & Noisy action & Action target & Visual target & Losses \\
\midrule
RGB
& Yes
& Yes
& Future RGB
& $\mathcal{L}_{\mathrm{act}}+\mathcal{L}_{\mathrm{rgb}}$ \\
Self-mask
& No
& No
& Future robot mask
& $\mathcal{L}_{\mathrm{mask}}$ \\
\bottomrule
\end{tabular}
}
\end{table}

\paragraph{Video prompts.}
For RGB instances, we use the prompt
\texttt{A video recorded from a robot's point of view executing the following instruction: {task}.}
For self-mask instances, we use
\texttt{A robot-arm mask video recorded from a robot's point of view executing the following instruction: {task}.}
Here, \texttt{{task}} is replaced with the natural-language task instruction from RoboTwin. A self-mask instance uses the corresponding initial RGB frame as its visual condition and predicts a sequence of binary robot-arm masks. Because the target includes only the robot arm, the prompt does not require an actor-color legend. The clean action condition is constructed using the standard RGB prompt and reused by the self-mask instance.

RGB and self-mask instances are sampled at a ratio of $9{:}1$. We set $\lambda_{\mathrm{act}}=\lambda_{\mathrm{video}}=1$ for the action and visual flow-matching losses.

\subsection{Temporal Sampling and Multi-View Preprocessing}
\label{sec:supp_temporal_alignment}

For a window starting at control step $t$, the policy target contains the 32 actions from $a_t$ to $a_{t+31}$. We collect the corresponding 32 future visual observations over the same horizon and downsample them to 8 frames in temporal order. The RGB and self-mask targets use the same frame offsets, so each mask frame is aligned with the matching RGB frame and the corresponding part of the action chunk.

\begin{figure*}[t]
\centering
\includegraphics[width=\textwidth]{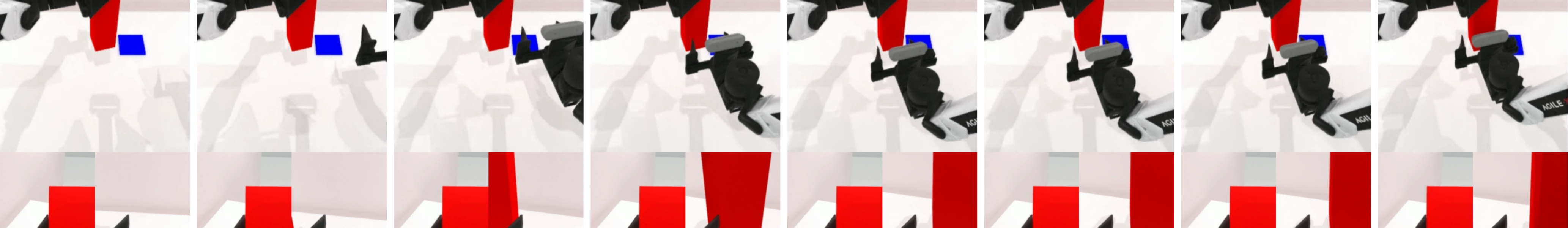}
\caption{\textbf{Temporal sampling and multi-view composition.} A representative 32-step future observation window after temporal downsampling to 8 frames, ordered from left to right. Each frame uses the fixed T-shaped composition of the three camera views. Future RGB and self-mask targets share the same temporal offsets and spatial layout.}
\label{fig:supp_temporal_sampling}
\end{figure*}

At each selected time step, the three camera views are placed in a fixed T-shaped layout and represented as a $384\times320$ composite. The view order and spatial placement are shared by the current RGB observation, future RGB target, and future self-mask target. For the physical robot, the three sources are one fixed high camera and the left and right wrist cameras. RoboTwin uses the corresponding three simulator views. Figure~\ref{fig:supp_temporal_sampling} illustrates the resulting eight-frame sequence and multi-view composition.

RGB and mask composites follow the same video encoding path, temporal positions, latent dimensions, and flow-matching noise schedule. The mask targets are binary robot self-masks and are used only as prediction targets; the policy input remains RGB in both instance types.

\subsection{Configuration}
\label{sec:supp_optimization}

Following \fastwam~\citep{yuan2026fastwam}, we initialize the shared video backbone from WAN2.2-5B~\citep{wan2025} and transfer its pretrained parameters to the action expert through weight interpolation. Parameters with matching shapes are copied directly, while semantically corresponding parameters with different shapes are resized to the action-expert dimensions. For each mismatched axis, we apply one-dimensional linear interpolation with aligned endpoints. When resizing an axis from $d_v$ to $d_a$, target index $j\in\{0,\ldots,d_a-1\}$ is mapped to the source coordinate
\[
u_j=\frac{j(d_v-1)}{d_a-1},
\]
and its value is obtained by linear interpolation between the two neighboring source entries. For tensors with multiple mismatched axes, the operation is applied sequentially along each axis. When the resized axis is the input dimension of a weight tensor, we additionally scale the interpolated weights by $\sqrt{d_v/d_a}$ to compensate for the change in input width and maintain comparable activation magnitudes. Action-specific input and output layers without corresponding video-backbone parameters are initialized separately. Table~\ref{tab:supp_implementation} summarizes the settings used in the reported experiments.

\begin{table}[!ht]
\centering
\caption{Optimization, inference, and system configuration.}
\label{tab:supp_implementation}
\setlength{\tabcolsep}{4pt}
\resizebox{\columnwidth}{!}{%
\begin{tabular}{@{}ll@{}}
\toprule
Setting & Value \\
\midrule
Video backbone & WAN2.2-5B \\
Input views & 3 \\
Composite resolution & $384\times320$ \\
Action horizon & 32 steps \\
Visual target length & 8 frames \\
Optimizer & AdamW \\
Learning rate & $1\times10^{-4}$ \\
Training epochs & 5 \\
$\lambda_{\mathrm{act}}, \lambda_{\mathrm{video}}$ & $1, 1$ \\
RGB:self-mask sampling ratio & $9{:}1$ \\
Global batch size & 1024 \\
Per-GPU batch size & 16 \\
Action denoising steps & 10 \\
Executed actions before replanning & 24 \\
Training hardware & 64 NVIDIA A800 GPUs \\
Training time & Approximately 30 hours \\
Operating system & Ubuntu 22.04 \\
Python / PyTorch & 3.10.12 / 2.7.1 \\
CUDA & 12.8 \\
System memory & 1 TB \\
\bottomrule
\end{tabular}%
}
\end{table}

During deployment, the policy encodes the latest observation, predicts a 32-step action chunk with 10 denoising steps, executes the first 24 actions, and then replans. The clean-action and future-visual streams are not instantiated. Optional visual rollout instead inserts a candidate action through the clean-action path and selects future RGB or future self-mask prediction with the corresponding output prompt.

\subsection{Datasets and Splits}

For RoboTwin 2.0, every method is trained on the same multi-task dataset of 27,500 demonstrations. This dataset contains 2,500 demonstrations from clean environments and 25,000 demonstrations collected under the benchmark's scene randomization protocol. The future-video analysis uses a uniformly sampled subset of 275 trajectories, corresponding to 1\% of this offline dataset. This analysis measures fidelity to demonstrated futures under matched context and action conditions; it should not be interpreted as a separate test of held-out video generalization.

The physical-robot dataset contains 17,307 episodes and 9,557,419 transitions. We use a fixed episode-level random split with seed 42. The training split contains 17,133 episodes and 9,461,919 transitions, while the validation split contains 174 episodes and 95,500 transitions. All real robot baselines are trained on the same training split. The validation episodes are excluded from gradient updates.

\section{Extended RoboTwin Evaluation}
\label{sec:supp_robotwin}

\subsection{Evaluation Protocol}

We evaluate 50 RoboTwin 2.0 tasks in both clean and randomized environments. Each method is evaluated for 100 simulator rollouts per task and setting, giving 5,000 evaluation episodes for each setting. We use the simulator's task success signal and report the unweighted mean success rate across tasks. Clean and randomized results are reported separately because the latter changes scene appearance and configuration while preserving the task objective.

\subsection{Task-Level Results}

Table~\ref{tab:task_level_results} reports the complete task-level results for the methods included in our per-task evaluation export. Relative to \fastwam, \ours achieves higher mean success rates in both clean and randomized environments, with a more pronounced improvement under randomization. The largest randomized gains occur on Open Microwave, Hanging Mug, Handover Block, and Place Mouse Pad. The gains are not uniform across all tasks, which is consistent with the modest aggregate improvement reported in the main paper. The overall pattern indicates that self-grounded supervision is most useful on a subset of tasks requiring stronger spatial alignment between visible robot motion and the manipulated scene.

\subsection{Future-Video Fidelity Protocol}
\label{sec:supp_video_fidelity}

We evaluate the observation-conditioned \fastwam objective and the clean-action-conditioned \ours objective. For every trajectory, future predictions are generated across its full duration using matched context windows. The two models use the same observation, instruction, target trajectory, and evaluation subset; the only intended change is whether the future visual stream receives the clean demonstrated action.

We report LPIPS \citep{zhang2018lpips}, PSNR \citep{huynh2008scope}, and FVD with I3D features \citep{unterthiner2018towards}. These reference-based metrics measure similarity to the recorded future, but they do not establish whether a model responds to controlled changes in the conditioning action. We therefore pair them with the action sensitivity analyses reported below.

\begin{table}[htbp]
\centering
\caption{WorldArena-based action sensitivity and raw consistency scores. Higher is better for all reported metrics.}
\label{tab:worldarena_ablation}
\scriptsize
\resizebox{\columnwidth}{!}{%
\begin{tabular}{@{}lcccc@{}}
\toprule
Model
& \shortstack{Action\\Following$\uparrow$}
& \shortstack{Subject\\Raw$\uparrow$}
& \shortstack{Background\\Raw$\uparrow$}
& \shortstack{Photometric\\Raw$\uparrow$} \\
\midrule
\fastwam
& 0.0000
& 0.9138
& 0.9309
& 0.8126 \\
\ours{} w/o self-mask
& 0.0181
& 0.9321
& 0.9380
& \textbf{1.5168} \\
\textbf{\ours}
& \textbf{0.0184}
& \textbf{0.9326}
& \textbf{0.9385}
& 1.4361 \\
\bottomrule
\end{tabular}%
}
\end{table}

\onecolumn
\begin{table}[t]
\centering
\caption{Task-level success rates (\%) on RoboTwin 2.0 under the clean and randomized settings.}
\label{tab:task_level_results}

\begingroup

\fontsize{9pt}{9pt}\selectfont

\renewcommand{\arraystretch}{1.12}

\begin{tabular}{@{}lcccccccccc@{}}
\toprule
\multirow{2}{*}{\centering Task}
& \multicolumn{2}{c}{SelfWAM}
& \multicolumn{2}{c}{w/o Mask}
& \multicolumn{2}{c}{FastWAM}
& \multicolumn{2}{c}{$\pi_{0.5}$}
& \multicolumn{2}{c}{Motus} \\
\cmidrule(lr){2-3}
\cmidrule(lr){4-5}
\cmidrule(lr){6-7}
\cmidrule(lr){8-9}
\cmidrule(lr){10-11}
& Clean & Rand.
& Clean & Rand.
& Clean & Rand.
& Clean & Rand.
& Clean & Rand. \\
\midrule
Adjust Bottle              & 100 & 100 & 100 & 99  & 100 & 99  & 100 & 99  & 89  & 93  \\
Beat Block Hammer          & 100 & 99  & 100 & 94  & 98  & 97  & 96  & 93  & 95  & 88  \\
Blocks Ranking RGB         & 100 & 97  & 98  & 95  & 100 & 99  & 92  & 85  & 99  & 97  \\
Blocks Ranking Size        & 98  & 98  & 89  & 92  & 93  & 96  & 49  & 26  & 75  & 63  \\
Click Alarmclock           & 100 & 100 & 100 & 100 & 100 & 100 & 98  & 89  & 100 & 100 \\
Click Bell                 & 100 & 100 & 100 & 100 & 100 & 100 & 99  & 66  & 100 & 100 \\
Dump Bin Bigbin            & 95  & 91  & 95  & 98  & 98  & 98  & 92  & 97  & 95  & 91  \\
Grab Roller                & 100 & 100 & 100 & 100 & 100 & 100 & 100 & 100 & 100 & 100 \\
Handover Block             & 93  & 93  & 88  & 77  & 96  & 84  & 66  & 57  & 86  & 73  \\
Handover Mic               & 97  & 100 & 98  & 99  & 100 & 98  & 98  & 97  & 78  & 63  \\
Hanging Mug                & 61  & 65  & 57  & 48  & 48  & 53  & 18  & 17  & 38  & 38  \\
Lift Pot                   & 99  & 100 & 100 & 99  & 100 & 100 & 96  & 85  & 96  & 99  \\
Move Can Pot               & 91  & 98  & 79  & 88  & 90  & 96  & 51  & 55  & 34  & 74  \\
Move Pillbottle Pad        & 99  & 96  & 99  & 98  & 100 & 100 & 84  & 61  & 93  & 96  \\
Move Playingcard Away      & 99  & 100 & 100 & 100 & 100 & 99  & 96  & 84  & 100 & 96  \\
Move Stapler Pad           & 73  & 75  & 79  & 72  & 72  & 69  & 56  & 42  & 83  & 85  \\
Open Laptop                & 98  & 98  & 100 & 100 & 97  & 97  & 90  & 96  & 95  & 91  \\
Open Microwave             & 74  & 67  & 68  & 72  & 54  & 41  & 34  & 77  & 95  & 91  \\
Pick Diverse Bottles       & 77  & 87  & 80  & 86  & 88  & 84  & 81  & 71  & 90  & 91  \\
Pick Dual Bottles          & 89  & 97  & 99  & 93  & 98  & 96  & 93  & 63  & 96  & 90  \\
Place A2B Left             & 96  & 99  & 89  & 97  & 94  & 96  & 87  & 82  & 88  & 79  \\
Place A2B Right            & 97  & 100 & 95  & 96  & 97  & 97  & 87  & 84  & 91  & 87  \\
Place Bread Basket         & 91  & 95  & 95  & 96  & 92  & 95  & 77  & 64  & 91  & 94  \\
Place Bread Skillet        & 87  & 95  & 95  & 93  & 94  & 96  & 85  & 66  & 86  & 83  \\
Place Burger Fries         & 99  & 99  & 93  & 99  & 98  & 99  & 94  & 87  & 98  & 98  \\
Place Can Basket           & 68  & 63  & 71  & 55  & 70  & 67  & 62  & 62  & 81  & 76  \\
Place Cans Plasticbox      & 98  & 97  & 100 & 97  & 99  & 97  & 94  & 84  & 98  & 94  \\
Place Container Plate      & 100 & 99  & 98  & 96  & 98  & 98  & 99  & 95  & 98  & 99  \\
Place Dual Shoes           & 95  & 94  & 86  & 89  & 86  & 91  & 75  & 75  & 93  & 87  \\
Place Empty Cup            & 100 & 100 & 98  & 100 & 100 & 98  & 100 & 99  & 99  & 98  \\
Place Fan                  & 93  & 96  & 97  & 96  & 97  & 94  & 87  & 85  & 91  & 87  \\
Place Mouse Pad            & 94  & 94  & 86  & 85  & 86  & 85  & 60  & 39  & 66  & 68  \\
Place Object Basket        & 88  & 86  & 80  & 76  & 87  & 86  & 80  & 76  & 81  & 87  \\
Place Object Scale         & 96  & 98  & 89  & 95  & 93  & 99  & 86  & 80  & 88  & 85  \\
Place Object Stand         & 97  & 93  & 95  & 87  & 96  & 96  & 91  & 85  & 98  & 97  \\
Place Phone Stand          & 97  & 95  & 97  & 98  & 99  & 98  & 81  & 81  & 87  & 86  \\
Place Shoe                 & 95  & 100 & 95  & 96  & 97  & 99  & 92  & 93  & 99  & 97  \\
Press Stapler              & 95  & 97  & 89  & 94  & 90  & 91  & 87  & 83  & 93  & 98  \\
Put Bottles Dustbin        & 96  & 89  & 87  & 87  & 92  & 95  & 84  & 79  & 81  & 79  \\
Put Object Cabinet         & 89  & 90  & 87  & 91  & 90  & 95  & 80  & 79  & 88  & 71  \\
Rotate QRcode              & 94  & 85  & 86  & 88  & 89  & 90  & 89  & 87  & 89  & 73  \\
Scan Object                & 92  & 91  & 87  & 87  & 93  & 93  & 72  & 65  & 67  & 66  \\
Shake Bottle               & 100 & 100 & 100 & 100 & 100 & 100 & 99  & 97  & 100 & 97  \\
Shake Bottle Horizontally  & 100 & 100 & 100 & 100 & 100 & 100 & 99  & 99  & 100 & 98  \\
Stack Blocks Three         & 98  & 95  & 97  & 96  & 98  & 96  & 91  & 76  & 91  & 95  \\
Stack Blocks Two           & 100 & 100 & 99  & 99  & 100 & 100 & 97  & 100 & 100 & 98  \\
Stack Bowls Three          & 73  & 81  & 82  & 83  & 77  & 78  & 77  & 71  & 79  & 87  \\
Stack Bowls Two            & 91  & 94  & 94  & 93  & 93  & 96  & 95  & 96  & 98  & 98  \\
Stamp Seal                 & 80  & 95  & 80  & 88  & 85  & 93  & 79  & 55  & 93  & 92  \\
Turn Switch                & 66  & 73  & 64  & 73  & 69  & 69  & 62  & 54  & 84  & 78  \\
\midrule
\textbf{Average}
& \textbf{92.16} & \textbf{93.08}
& 90.80 & 90.80
& 91.82 & 91.86
& 82.74 & 76.76
& 88.66 & 87.02 \\
\bottomrule
\end{tabular}

\endgroup
\end{table}
\twocolumn

\section{Action Sensitivity and Controlled Rollouts}
\label{sec:supp_action_sensitivity}

\subsection{WorldArena Metrics}

Standard reference-based metrics measure how closely a predicted future matches the ground-truth video, but do not reveal whether the prediction responds to the supplied action. We therefore adopt Action Following and three consistency metrics from WorldArena~\citep{2026worldarena}. For each context, we fix the initial observation and sampling noise while varying only the conditioning action chunk.

Given $M$ generated futures and their global CLIP features $\{f_i\}_{i=1}^{M}$, Action Following is the average pairwise feature dissimilarity
\begin{equation}
S_{\mathrm{act}}
=
\frac{2}{M(M-1)}
\sum_{i<j}
\left(
1-
\frac{f_i^\top f_j}{\lVert f_i\rVert\lVert f_j\rVert}
\right).
\label{eq:supp_action_following}
\end{equation}
A larger value indicates that the model produces more distinct futures when the conditioning action changes. This metric diagnoses action sensitivity rather than physical correctness.

For a generated video $V=\{I_t\}_{t=1}^{T}$, Subject Raw Consistency uses DINO features $d_t$, and Background Raw Consistency uses CLIP image features $c_t$:
\begin{equation}
S_{\mathrm{subj}}^{\mathrm{raw}}
=
\frac{1}{T-1}
\sum_{t=2}^{T}
\frac{\cos(d_t,d_1)+\cos(d_t,d_{t-1})}{2},
\label{eq:supp_subject_raw}
\end{equation}
\begin{equation}
S_{\mathrm{bg}}^{\mathrm{raw}}
=
\frac{1}{T-1}
\sum_{t=2}^{T}
\frac{\cos(c_t,c_1)+\cos(c_t,c_{t-1})}{2}.
\label{eq:supp_background_raw}
\end{equation}
Photometric Raw Consistency is based on forward and backward optical-flow warping. Let $E_{\mathrm{photo}}$ be the mean round-trip endpoint error and $S_{\mathrm{dyn}}$ the WorldArena dynamic-degree score. We report the pre-normalized quantity
\begin{equation}
S_{\mathrm{photo}}^{\mathrm{raw}}
=
\frac{1}{E_{\mathrm{photo}}}
\min\left(1,\frac{S_{\mathrm{dyn}}}{\gamma}\right),
\label{eq:supp_photometric_raw}
\end{equation}
where $\gamma$ is the benchmark's motion threshold. Because this is a raw reciprocal error score, it is not restricted to $[0,1]$. We report raw consistency values to avoid dependence on empirical normalization bounds from a different model pool. High consistency can still result from nearly static predictions, so all three consistency metrics are interpreted jointly with Action Following and the controlled rollouts below.

Table~\ref{tab:worldarena_ablation} shows that clean-action conditioning accounts for most of the improvement in action sensitivity. Adding self-mask supervision further improves subject and background consistency, while photometric consistency decreases slightly. Overall, the complete model preserves strong action sensitivity while achieving broader improvements in visual consistency.

\subsection{Directional Action Perturbation} 

To construct directional action variants, we perturb the demonstrated action along the vertical and horizontal translation axes, producing upward, downward, leftward, and rightward variants. The perturbation magnitude increases linearly over the first eight action steps and remains fixed for the rest of the chunk. We keep the observation, instruction, and sampling noise fixed across variants to isolate the effect of action conditioning. The qualitative results in the main paper show that the action-unconditioned baseline remains nearly invariant, whereas \ours shifts the predicted robot self-mask in the perturbed direction and follows the corresponding simulator motion trend.

\subsection{Continuous Action Interpolation}

We additionally test whether the learned response changes continuously between a stationary action and the demonstrated action. Let $\mathbf a^{\mathrm{hold}}$ denote the hold action and $\mathbf a^{\mathrm{demo}}$ the demonstrated action. We construct
\begin{equation}
\mathbf a(\alpha)
=
(1-\alpha)\mathbf a^{\mathrm{hold}}
+
\alpha\mathbf a^{\mathrm{demo}},
\qquad
\alpha\in[0,1].
\label{eq:supp_action_interpolation}
\end{equation}
The visualization uses $\alpha\in\{0,0.25,0.75,1\}$ and keeps the context and sampling noise fixed.

\begin{figure}[t]
\centering
\includegraphics[width=\columnwidth]{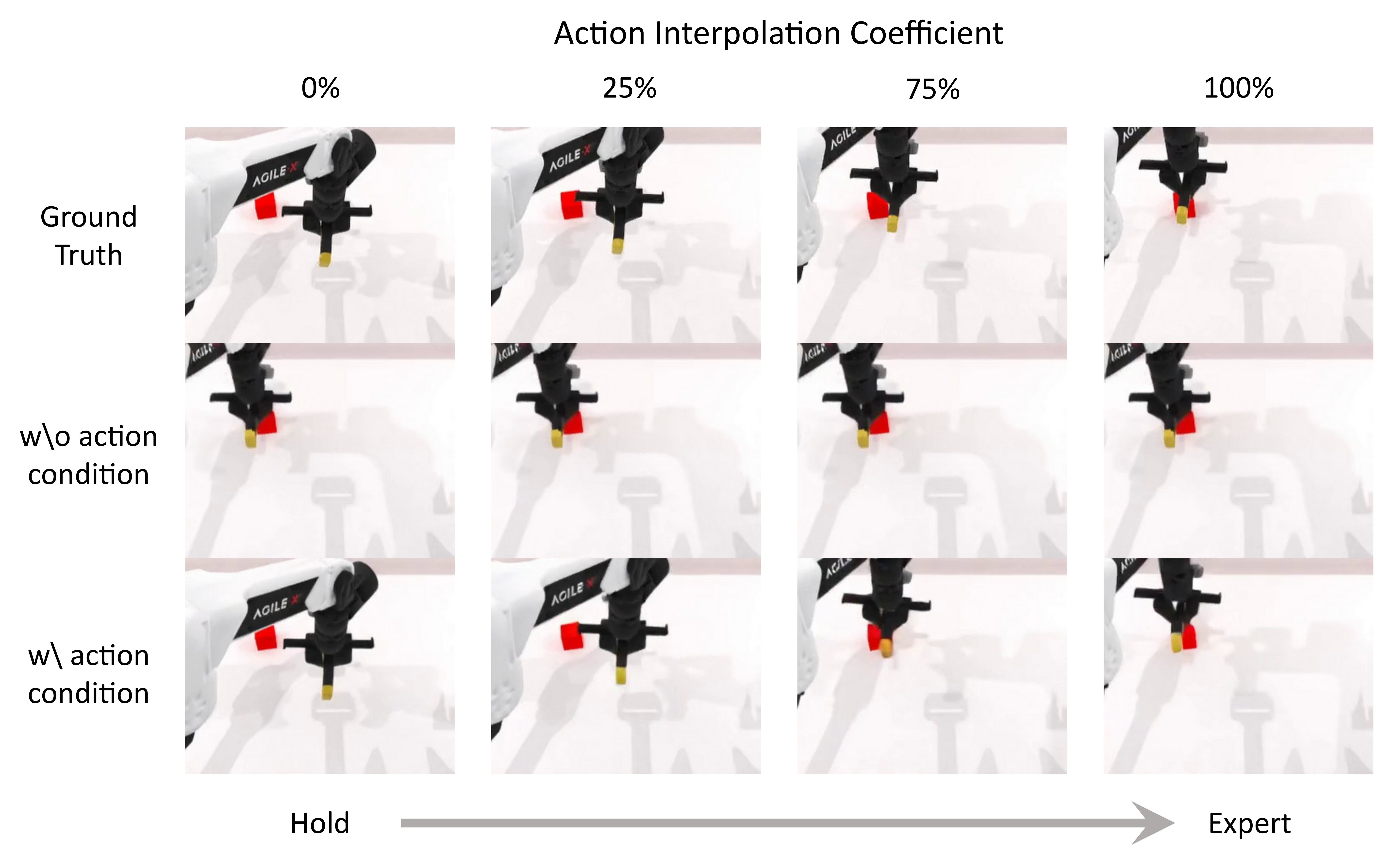}
\caption{\textbf{Continuous action interpolation.} We interpolate from a stationary hold action ($0\%$) to the demonstrated action ($100\%$). The action-unconditioned baseline produces nearly invariant futures, whereas \ours changes progressively toward the demonstrated-action outcome.}
\label{fig:action_interpolation}
\end{figure}

\begin{figure*}[t]
\centering
\includegraphics[width=0.95\textwidth]{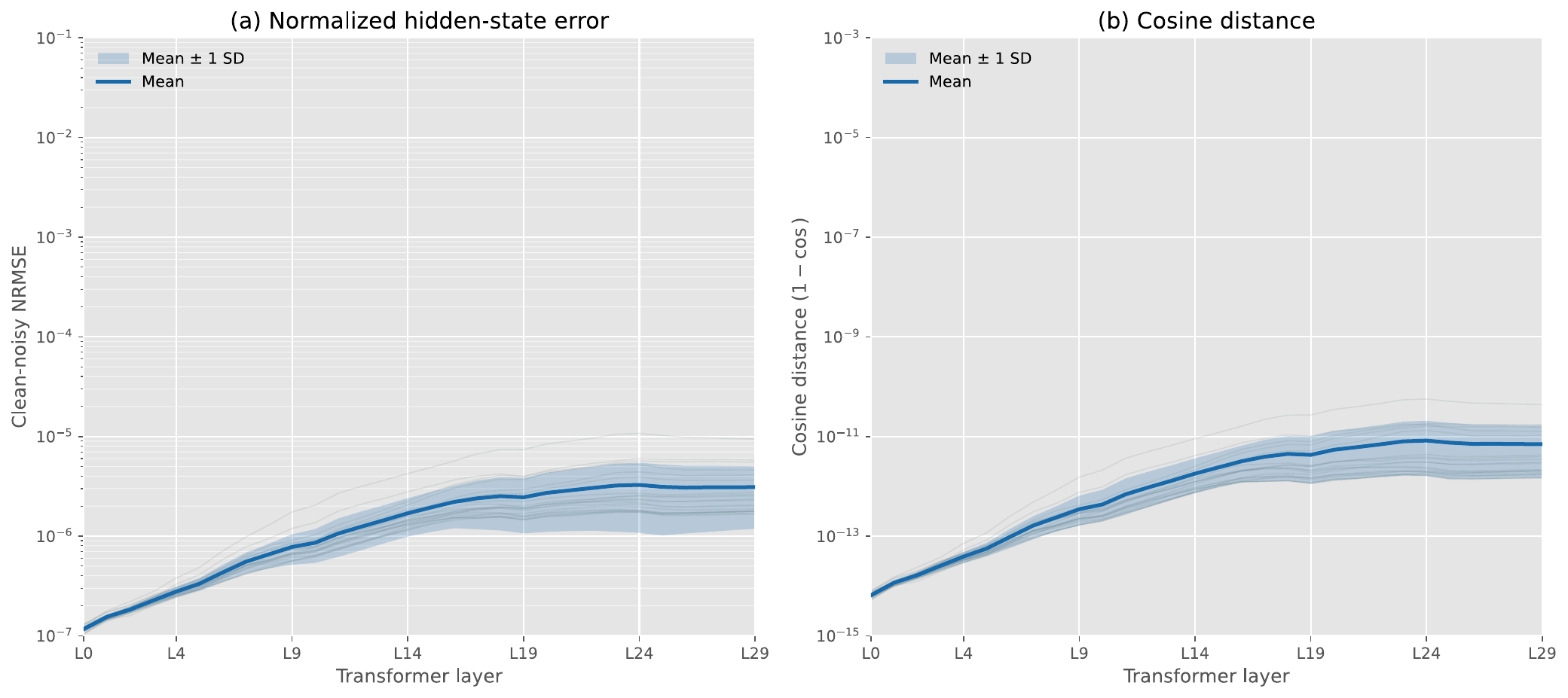}
\caption{\textbf{Consistency of the clean and noisy action paths at $\sigma=0$.}
We compare aligned action representations from the two paths at every Transformer layer using (a) NRMSE and (b) cosine distance. Solid curves and shaded regions show the mean and $\pm1$ standard deviation, respectively.}
\label{fig:exact_endpoint_consistency}
\end{figure*}

As shown in Figure~\ref{fig:action_interpolation}, the action-unconditioned baseline predicts nearly unchanged RGB futures as $\alpha$ increases. In contrast, \ours progresses from the hold outcome toward the demonstrated-action outcome in the same qualitative direction as the corresponding rollout. The metric results, directional perturbations, and interpolation study provide complementary evidence that the learned future prediction is action dependent, directionally aligned, and locally continuous around demonstrated actions.

\section{Consistency of the Clean and Noisy Action Paths}
\label{sec:supp_endpoint_consistency}

Our training design represents each demonstrated action through two paths: a clean copy that conditions future visual prediction and a noise-perturbed copy used for action denoising. The two paths share the same action-side network, but their tokens serve different roles under the directed attention pattern. We therefore examine whether they produce consistent representations when the perturbation on the noisy path is removed.

Let $h_l^{\mathrm{clean}}$ and $h_l^{\mathrm{noisy}}(\sigma)$ denote the aligned action-token representations after Transformer layer $l$ on the clean and noisy paths, respectively. At $\sigma=0$, the noisy path receives the original demonstrated action without perturbation. We compare $h_l^{\mathrm{clean}}$ and $h_l^{\mathrm{noisy}}(0)$ at every Transformer layer using normalized root mean squared error (NRMSE) and cosine distance. Specifically, we define
\[
\operatorname{NRMSE}(x,y)
=
\frac{
\sqrt{\frac{1}{N}\sum_{i=1}^{N}(x_i-y_i)^2}
}{
\sqrt{\frac{1}{N}\sum_{i=1}^{N}x_i^2}+\epsilon
},
\]
where $x$ is the clean-path representation, $y$ is the corresponding noisy-path representation, and $N$ is the number of aligned token-channel elements. Both metrics are computed over aligned action tokens and feature channels.

As shown in Figure~\ref{fig:exact_endpoint_consistency}, the discrepancies remain small throughout the action network. They increase gradually with network depth, with no abrupt increase at any individual layer. Across the sampled validation windows and all Transformer layers, the maximum NRMSE and cosine distance are approximately $1\times10^{-5}$ and $6\times10^{-11}$, respectively. These results indicate that the clean and noisy paths produce numerically consistent representations when evaluated on the same unperturbed action. In particular, we observe no layer-specific mismatch attributable to their different roles in the directed attention pattern. The clean action representation can therefore be viewed as the zero-noise endpoint of the action denoising path.

\section{Real Robot Experimental Details}
\label{sec:supp_real_robot}

\subsection{Platform and Data}

We conduct physical experiments on a stationary AgileX ALOHA dual-arm platform \citep{agilexAlohaConfig}. The full teleoperation setup pairs the two follower arms with two leader arms for data collection. Visual observations are captured by two wrist cameras and one fixed high camera, following the standard Cobot Magic camera configuration \citep{agilexCobotMagic}. No mobile-base command is included during training or evaluation.

The policy uses joint-position control. Low-level robot commands are executed at 30 Hz. Each action contains six target joint angles and one gripper command for each arm, giving a 14-dimensional action vector; proprioception follows the same 14-dimensional ordering \citep{agilexAlohaConfig}. 

\subsection{RobotSeg Mask Generation}
\label{sec:supp_robotseg}

RobotSeg is a robot-specific image and video segmentation model built on SAM 2 \citep{mei2026robotseg}. It introduces a Robot Prompt Generator that initializes segmentation from semantic robot categories without manual spatial prompts, together with a Structure-Enhanced Memory Associator that preserves articulated robot structure during temporal propagation. RobotSeg supports arm, gripper, and whole-robot categories; we use the whole-robot category for all physical-robot targets.

\paragraph{Robot mask generation.}
We generate robot-arm masks offline using the released RobotSeg checkpoint \citep{robotsegCode2026}. Each camera stream is processed independently using the automatic \texttt{robot} category, without manual points, bounding boxes, or reference masks. The resulting binary masks are arranged into the same T-shaped layout as the RGB observations and used only as future self-mask targets. The policy receives no mask input, and RobotSeg is not used during deployment.

\subsection{Real-Robot Evaluation Details}

We use one fixed language instruction for each task:
\begin{enumerate}
\item Pick up the brown cup and place it on the wooden stand.
\item Pick up the black pen and place it in the blue cup.
\item Pick up the black mouse and place it on the gray mouse pad.
\item Pick up the paper ball and place it in the desktop bin.
\end{enumerate}

For each trial, we randomly place both the manipulated object and the target within predefined regions of the workspace. The policy may make multiple attempts within the same episode, such as regrasping the object or correcting an unsuccessful placement, without resetting the scene. An episode ends when the task is completed or when the current scene state no longer allows the policy to complete the task without human intervention.

\section{Limitations and Future Work}

\label{sec:supp_limitations}

SelfWAM predicts future RGB observations and robot masks, but does not explicitly model contact, force, or interactions hidden by occlusion. Extending the prediction targets with these signals could provide richer supervision for contact-rich manipulation and partially observed interactions.

% Use a single bibliography for both the main paper and supplementary material.
\bibliography{SelfWAM_AAAI2027_Draft}

\end{document}